\documentclass{article}
\usepackage{arxiv}
\usepackage[utf8]{inputenc}
\usepackage[printonlyused,nohyperlinks]{acronym} 
\usepackage{amsmath}
\usepackage{amssymb}
\usepackage{bm}
\usepackage[T1]{fontenc}
\usepackage{amsthm}
\usepackage{color, boldline}
\usepackage{hyperref}
\usepackage{graphicx}
\usepackage{tikz}
\usepackage{todonotes}
\usepackage{booktabs}
\usepackage{tabularx}
\usetikzlibrary{shapes,arrows,fit}

\DeclareMathOperator*{\argmax}{arg\,max}

\usepackage{xcolor}
\definecolor{darkblue}{rgb}{0, 0, 0.5}
\hypersetup{colorlinks=true,citecolor=darkblue, linkcolor=darkblue, urlcolor=darkblue}

\usepackage{natbib}
\title{Optimizing small BERTs trained for German \acs{NER}}
\author{\href{https://orcid.org/0000-0002-3889-6629}{\includegraphics[scale=0.06]{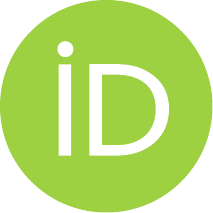}\hspace{1mm}Jochen Zöllner} \\
    University of Rostock \\
    Institute of Mathematics \\
	\texttt{jochen.zoellner@uni-rostock.de} \\
	\And
	\href{https://orcid.org/0000-0003-3856-5878}{\includegraphics[scale=0.06]{orcid.pdf}\hspace{1mm}Konrad Sperfeld} \\
	University of Rostock \\ 
	Institute of Mathematics \\ 
	\texttt{konrad.sperfeld@uni-rostock.de} \\
	\And
	\href{https://orcid.org/0000-0003-3958-6240}{\includegraphics[scale=0.06]{orcid.pdf}\hspace{1mm}Christoph Wick} \\
	Planet AI GmbH Rostock \\ 
	\texttt{christoph.wick@planet-ai.de} \\
	\And
	\href{https://orcid.org/0000-0003-1901-9644}{\includegraphics[scale=0.06]{orcid.pdf}\hspace{1mm}Roger Labahn} \\
	University of Rostock \\
	Institute of Mathematics \\ 
	\texttt{roger.labahn@uni-rostock.de} \\
}

\begin{document}
\title{Optimizing small BERTs trained for German \acs{NER}}






\maketitle

\begin{abstract}
Currently, the most widespread neural network architecture for training language models is the so called \acs{BERT} which led to improvements in various \ac{NLP} tasks.
In general, the larger the number of parameters in a \acs{BERT} model, the better the results obtained in these \ac{NLP} tasks.
Unfortunately, the memory consumption and the training duration drastically increases with the size of these models. 
In this article, we investigate various training techniques of smaller \acs{BERT} models: 
We combine different methods from other \acs{BERT} variants like ALBERT, RoBERTa, and relative positional encoding.
In addition, we propose two new fine-tuning modifications leading to better performance: \acl{CSE} tagging and a modified form of \aclp{LCRF}. Furthermore, we introduce \acl{WWA} which reduces \acsp{BERT} memory usage and leads to a small increase in performance compared to classical Multi-Head-Attention. 
We evaluate these techniques on five public German \ac{NER} tasks of which two are introduced by this article.

\end{abstract}
\acresetall

\section{Introduction}
\ac{NER} is a well-known task in the field of \ac{NLP}. 
The NEISS\footnote{\url{https://www.neiss.uni-rostock.de}} 
project in which we work in close cooperation with Germanists is devoted to the automation of diverse processes during the creation of digital editions.
One key task in this area is the automatic detection of entities in text corpora which corresponds to a common \ac{NER} task.
Currently, the best results for \ac{NER} tasks have been achieved with Transformer-based~\citep{vaswani2017attention} language models such as \acf{BERT}~\citep{devlin2018bert}.
Classically, a \ac{BERT} is first pre-trained with large amounts of unlabeled text to obtain a robust language model and then fine-tuned to a downstream task.
Especially for the pre-training step, many variants of \ac{BERT} like ALBERT~\citep{lan2019albert}, RoBERTa~\citep{liu2019roberta}, or XLNet~\citep{yang2019xlnet} were already investigated.
Pre-training is resource-intensive and takes much time (several weeks) for training. For that reason online-platforms such as Hugging Face\footnote{\url{https://huggingface.co/}} offer a zoo of already pre-trained networks that can directly be used to train a downstream task.
However, the available models are not always suitable for a certain task such as \ac{NER} in German because they can be pre-trained on a different domain (e.g., language, time epoch, or text style).

Furthermore, when philologists create new digital editions, different research priorities can be set, so that a different associated \ac{NER} task is created each time. That is why philologists must also be able to train individual \ac{NER} tasks themselves who commonly only have access to limited compute resources. For this reason, the aim is to train \ac{NER} tasks on smaller \ac{BERT} models as best as possible. 
Since our focus is the  \ac{NER}, we test if the optimizations work consistently on five different German \ac{NER} tasks. Due to our project aims we evaluated our new methods on the German language. We suspect a consistent behavior on similar European languages like English, French and Spanish.
Two of the considered tasks rely on new \ac{NER} datasets which we generated from existing digital text editions.

Therefore, in this article, we examine which techniques are optimal to pre-train and fine-tune a \ac{BERT} to solve \ac{NER} tasks in German with limited resources.
We investigate this on smaller \ac{BERT} models with six layers that can be pre-trained on a single GPU (RTX 2080 Ti 11\,GB) within 9 days while fine-tuning can be performed on a notebook CPU in a few hours.

We first compared different well-established pre-training techniques such as \acf{MLM}, \acf{SOP}, and \acf{NSP} on the final result of the downstream \ac{NER} task.
Furthermore, we investigated the influence of absolute and relative positional encoding, as well as \acf{WWM}.

As a second step, we compared various approaches for carrying out fine-tuning, since the tagging rules cannot be learned consistently by classical fine-tuning approaches.
In addition to existing approaches such as the use of \acfp{LCRF}, we propose the so-called \ac{CSE} tagging and an specially modified form of \acp{LCRF} for \ac{NER} which led to an increased performance.
Furthermore, for decoding, we introduced a simple rule-based approach, which we call Entity-Fix rule, to further improve the results.

As already mentioned, the training of a \ac{BERT} requires many resources.
One of the reasons is that the memory amount of \ac{BERT} depends quadratically on the sequence length when calculating the energy values (attention scores) in its attention layers which leads to memory problems for long sequences. 
In this article, we propose \acl{WWA}, a new modification of the Transformer architecture that not only reduces the number of energy values to be calculated by about factor two, but also results in slightly improved results.

In summary, the main goal of this article is to enable the training of efficient \ac{BERT} models for German \ac{NER} on limited resources.
For this, the article provides different methodology and claims the following contributions:
\begin{itemize}
    \item We introduce and share two datasets for German \ac{NER} formed from existing digital editions.
    \item We investigate the influence of different \ac{BERT} pre-training methods, such as pre-training tasks, varying positional encoding, and adding \acl{WWM} on a total of five different \ac{NER} datasets.
    \item On the same \ac{NER} tasks, we investigate different approaches to perform fine-tuning.
    Hereby, we propose two new methods which led to performance improvements: \acl{CSE} tagging and a modified form of \aclp{LCRF}.
    \item We introduce a novel rule-based decoding strategy achieving further improvements.
    \item We propose \acl{WWA}, a modification of the \ac{BERT} architecture that reduces the memory requirements of the \ac{BERT} models, especially for processing long sequences, and also leads to further performance improvements.
    \item We share the datasets (see Section \ref{sec:Datasets}) and our source code\footnote{\url{https://github.com/NEISSproject/tf2_neiss_nlp/tree/berNer21} which is based on \emph{tfaip} \citep{Wick2021}} with the community.
\end{itemize}

The remainder of this article is structured as follows:
In Section \ref{sec:Datasets} we present our datasets including the two new German \ac{NER} datasets.
In Section \ref{sec:Pre-training} we introduce the different pre-training techniques, while Section \ref{sec:Fine-tuning} describes fine-tuning.
Subsequently, in Section \ref{sec:WWA}, we introduce \acf{WWA}.
In all these sections we provide an overview of the existing techniques with the corresponding related work which we adopted and also introduce our novel methods.
After that, Section \ref{sec:Experiments} shows the conducted experiments  and their results.
We conclude this article by a discussion of our results and giving an outlook on future work.

\section{Datasets}\label{sec:Datasets}

In this section, we list the different datasets.
First, we describe the dataset used for pre-training throughout our experiments. Then, we mention the key attributes of five \ac{NER} datasets for the downstream tasks.

\subsection{Pre-training Data}
To pre-train a \ac{BERT}, a large amount of unlabeled text is necessary as input data.
We collected the German Wikipedia and a web crawl of various German newspaper portals to pre-train our \ac{BERT}. 
The dump of the German Wikipedia was preprocessed by the Wiki-Extractor~\citep{Wikiextractor2015} resulting in about 6\,GB of text data.
In addition, we took another 2\,GB of German text data from different newspaper portals\footnote{We used various German newspaper portals like \url{https://www.faz.net/aktuell/} or \url{https://www.berliner-zeitung.de/}, on August 2020} crawled with the news-please framework~\citep{Hamborg2017} .

\subsection{NER Downstream-Datasets}\label{sec:NERtasks}
We evaluated our methods on five different \ac{NER} tasks.
In addition to three already existing German \ac{NER} datasets, the frequently used GermEval 2014 dataset and two \ac{NER} datasets on German legal texts, we introduce two \ac{NER} tasks of two existing digital editions.
In the following, we describe each of the five tasks.

\paragraph{GermEval 2014}
One of the most widespread German \ac{NER} datasets is GermEval 2014 \citep{BenikovaBiemannKisselewetal2014} which comprises several News Corpora and Wikipedia.
In total, it contains about 590,000 tokens with about 41,000 entities which are tagged into four main entity classes: ``person'', ``organisation'', ``location'', and ``other''.
Each main class can appear in a default, a partial, or a derived variant, resulting in 12 overall classes.
In the GermEval task, entities can be tagged in two levels: outer and inner (nested entities). Since there are few inner annotations in the dataset, we restrict ourselves to evaluating the outer entities in our experiments as it is often the approach in other papers \citep[e.g. ][]{labusch_bert_2019,chan2020german,riedl_named_2018}. 
This is called the outer chunk evaluation scheme which is described in more detail by \cite{riedl_named_2018}.

\paragraph{\acl{LER}}
The \ac{LER} dataset \citep{leitner_dataset_2020} contains 2.15 million tokens with 54,000 manually annotated entities from German court decision documents of 2017 and 2018. 
The entities are divided into seven main classes and 19 subclasses which we label by Coarse-Grained (CG) and Fine-Grained (FG), respectively.
The FG task (\ac{LER} FG) is more difficult than the CG task (\ac{LER} CG)  due to its larger number of possible classes.

\paragraph{Digital Edition: Essays from H.\,Arendt}
We created an \ac{NER} dataset based on the digital edition ``Sechs Essays'' by H.\,Arendt.
It consists of 23 documents from the period 1932-1976 which are published  online in \citep{Arendt} as TEI files~\citep{TEI}. 
In these documents, certain entities were manually tagged.
Since some of the original \ac{NER} tags comprised too few examples and some ambiguities (e.g., place and country), we joined several tags as shown in Table~\ref{tab:arendt}.
\begin{table}[h]
\caption{Distribution of \ac{NER} entities in H.\,Arendt Edition. Column ``Original attributes'' lists which attributes from the original TEI files were combined into one ``Entity'' for the NER dataset. On average, an entity consists of 1.36 words.}
\label{tab:arendt}
\centering
    
	\vspace{5pt}
	\begin{tabular}{l|rrrr|p{3.1cm}}
	    \toprule
	    \textbf{Entity} & \multicolumn{1}{c}{\textbf{\# All}} & \multicolumn{1}{c}{\textbf{\# Train}} & \multicolumn{1}{c}{\textbf{\# Test}} & \multicolumn{1}{c|}{\textbf{\# Devel}} & \textbf{Original attributes} \\
	    \midrule 
	    person       & 1,702 & 1,303 & 182 & 217 & person, biblicalFigure, ficticiousPerson, deity, mythologicalFigure  \\ 
	    place        & 1,087 & 891 & 111 & 85 & place, country \\ 
	    ethnicity    & 1,093 & 867 & 115 & 111 & ethnicity  \\ 
	    organisation & 455 & 377 & 39 & 39 & organisation   \\ 
	    event        & 57 & 49 & 6 & 2 & event   \\ 
	    language     & 20 & 14 & 4 & 2 & language    \\ \midrule
	    unlabeled words & 153,223  & 121,154 & 16,101 & 15,968 & \\
	    \bottomrule
	    
	\end{tabular}
	
\end{table}
Note that we removed any annotation of the class ``ship'' since only four instances were available in the dataset and no other similar class is available.
We provide the resulting dataset online\footnote{\url{https://github.com/NEISSproject/NERDatasets/tree/main/Arendt}} in a format similar to the CONLL-X format~\citep{buchholz_conll-x_2006} and in a simple JSON format under a CC BY-NC-SA 3.0 DE license 
together with the training, development, and test partition. Since not all entities are equally distributed over the 23 documents, the sentences of all documents are shuffled before splitting them into partitions.

\paragraph{Digital Edition: Sturm Edition} The second \ac{NER} dataset consists of 174 letters of the years 1914-1922 from the Sturm Edition \citep{sturm_2018} available online in TEI format.
It is much simpler than the dataset from the H.\,Arendt edition and contains only persons, places, and dates as tagged entities. From the original TEI files, we built an \ac{NER} dataset with tags distributed as shown in Table~\ref{tab:sturm}.
Similarly to the H.\,Arendt dataset, the resulting dataset is available online\footnote{\url{https://github.com/NEISSproject/NERDatasets/tree/main/Sturm}} in a format similar to the CONLL-X format and in a simple JSON format under a CC-BY 4.0 license 
together with the training, development, and test partition. In contrast to the H.\,Arendt dataset, we split the 174 letters without shuffling the sentences across all documents.

\begin{table}[h]
	\caption{Distribution of \ac{NER} entities in the Sturm Edition. On average, an entity consists of 1.12 words.}
\label{tab:sturm}
\centering
    
	\vspace{5pt}
	\begin{tabular}{p{2.8cm}|rrrr}
	\toprule
	    \textbf{Entity} & \multicolumn{1}{c}{\textbf{\# All}} & \multicolumn{1}{c}{\textbf{\# Train}} & \multicolumn{1}{c}{\textbf{\# Test}} & \multicolumn{1}{c}{\textbf{\# Devel}} \\
	    \midrule
	    person       & 930 & 763 & 83 & 84 \\ 
	    date         & 722 & 612 & 59 & 51 \\ 
	    place        & 492 & 374 & 59 & 59    \\ \midrule
	    unlabeled words   & 33,809 & 27,047 & 3,306 & 3,456  \\
	   \bottomrule 
	\end{tabular}

\end{table}

\section{Pre-training Techniques}\label{sec:Pre-training}
In this section, we provide an overview of several common pre-training techniques for a \ac{BERT} which we examined in our experiments. 
\subsection{Pre-training Tasks}\label{sec:Pre-train}
In the original \ac{BERT} \citep{devlin2018bert}, pre-training is performed by simultaneously minimizing the loss of the so called \acf{MLM} and the \acf{NSP} task.
The \ac{MLM} task first tokenizes the text input with a subword tokenizer, then 15\% of the tokens are chosen randomly.
Hereby, 80\% of these chosen tokens are replaced by a special mask token, 10\% are replaced by a randomly chosen other token, and the remaining 10\% keep the original correct token. Therefore, the goal of the \ac{MLM} task is to find the original token for the 15\% randomly chosen tokens which is only possible by understanding the language and thus learning a robust language model.

Since \ac{BERT} should also be able to learn the semantics of different sentences within a text, \ac{NSP} was additionally included. When combining \ac{NSP} with \ac{MLM}, the input for pre-training are two masked sentences which are concatenated and separated by a special separator token.
In 50\% of the cases, two consecutive sentences from the same text document are used whereas in the other 50\% two random sentences from different documents are selected. 
The goal of the \ac{NSP} task is to identify which of the two variants it is.

In the follow-up papers RoBERTa~\citep{liu2019roberta} and XLNet~\citep{yang2019xlnet}, experiments showed that the \ac{NSP} task often had no positive effect on the performance of the downstream tasks.
Therefore, both papers recommended that the pre-training should solely be performed by the \ac{MLM} task.
In the ALBERT paper~\citep{lan2019albert} this was investigated in more detail. They assumed that the ineffectiveness of the \ac{NSP} task was only due to its simplicity which is why they introduced \acf{SOP} as a more challenging task that aims to learn relationships between sentences similar to the \ac{NSP} task: \ac{BERT} always receives two consecutive sentences, but in 50\% of the cases the order is wrong by flipping them.
The \ac{SOP} task is to learn the correct order of the sentences.

In this article, we examine the influences of the different pre-training tasks (\ac{MLM}, \ac{NSP}, \ac{SOP}) with the focus on improving the training of \ac{BERT} for German \ac{NER} tasks.

\subsection{Absolute and Relative Positional Encoding}
The original Transformer architecture~\citep{vaswani2017attention} was based exclusively on attention mechanisms to process input sequences. Attention mechanisms allow every sequence element to learn relations to all other elements. By default, Attention does not take into account information about the order of the elements in the sequence. But since information about the order of the input sequence elements is mandatory in almost every \ac{NLP} tasks, the original Transformer architecture introduced the so-called absolute positional encoding: a fixed position vector $p_j \in \mathbb{R}^{d_{model}}$ was added to each embedded input sequence element $x_j$ at position $j \in \left\{ 1,\ldots,n \right\}$ for an input sequence of length $n$, thus

\begin{equation*}
    x_j'=x_j+p_j.
\end{equation*}

In the original approach the position vector $p_j$ is built by computing sinusoids of different wavelength in the following way:

\begin{eqnarray*}
        p_{j,2k}&:=& \sin\left( j/10000^{2k/d_{model}}\right), \\
        p_{j,2k+1}&:=& \cos\left( j/10000^{2k/d_{model}}\right) 
\end{eqnarray*}
where $k \in \left\{ 1,\ldots,\lfloor \frac{d_{model}}{2}\rfloor \right\}$. While the experiments in \citep{vaswani2017attention} showed great results, the disadvantage of absolute positional encoding is that the performance is significantly reduced in cases where the models are applied on sequences longer than those on which they were trained because the respective position vectors were not yet seen during training.
Therefore, in \citep{rosendahl2019analysis} other variants for positional encoding were investigated and compared on translation tasks.
The most promising approach was relative positional encoding \citep{shaw2018self}: a trainable distance information $d_{j-i}^K$ is added in the attention layer when computing the energy $e_{i,j}$ of the $i$th sequence element to the $j$th one. Thus, if $x_i$ and $x_j$ are the $i$th and $j$th input elements of a sequence in an attention layer, instead of multiplying just the query vector $W^Qx_i$ with the key vector $W^Kx_j$, one adds the trainable distance information $d_{j-i}^K$ to the key vector resulting in

\begin{equation}\label{eq:attenergy}
	   e_{i,j}:=\frac{\left(W^Qx_i\right)^T\left(W^Kx_{j}+d_{j-i}^K\right)}{\sqrt{d_k}}
\end{equation}
where $W_n^Q,W_n^K \in \mathbb{R}^{d_{model}\times d_k}$. In addition, when multiplying the energy (after applying softmax) with the values, another trainable distance information $d_{j-i}^V$ is added. Finally, the output $y_i$ for the $i$th sequence element of a sequence of length $n$ with relative positional encoding is computed by

\begin{equation}\label{eq:RelAtt}
    y_i=\sum\limits_{j=1}^n\alpha_{i,j}\left(W^Vx_{j}+d_{j-i}^V\right)
\end{equation}
where $\alpha_{i,j}=\frac{\exp\left(e_{i,j}\right)}{\sum\limits_{k=1}^n\exp\left(e_{i,k}\right)}$ and $W^V\in\mathbb{R}^{d_{model}\times d_v}$. To train $d_{j-i}^K$ and $d_{j-i}^V$, a hyperparameter $\tau$ (called the clipping distance), the trainable embeddings $r_{-\tau}^K,\ldots,r_{\tau}^K\in\mathbb{R}^{d_k}$, and $r_{-\tau}^V,\ldots,r_{\tau}^V\in\mathbb{R}^{d_v}$ are introduced.
These embeddings are used to define the distance terms $d_{j-i}^K$ and $d_{j-i}^V$, where distances longer than the clipping distance $\tau$ are represented by $r_{\tau}$ or $r_{-\tau}$, thus:

\begin{eqnarray}
    d_{j'-j}^K & = & r_{\textrm{clip}_\tau(j'-j)}^K \label{eqn:distvectK} \\
    d_{j'-j}^V & = & r_{\textrm{clip}_\tau(j'-j)}^V \label{eqn:distvectV} \\
    \textrm{clip}_\tau(x) & = & \max\left( -\tau, \min\left(\tau,x\right)\right) \nonumber
\end{eqnarray}
\cite{rosendahl2019analysis} already showed that relative positional encoding suffers less from the disadvantages of absolute position encoding of unseen sequence lengths.
In this article, we examine the influence of these two variants of positional encoding during the training of German \ac{BERT} models.

\subsection{\acf{WWM}}
\acf{WWM} is a small modification of the \acf{MLM} task described in section~\ref{sec:Pre-train}. In contrast to the classic \ac{MLM} task, \ac{WWM} does not mask token-wisely but instead word-wisely. This means that in all cases either all tokens belonging to a word are masked or none of them. Recent work of \cite{chan2020german,cui2019pre} already showed the positive effect of \ac{WWM} in pre-training on the performance of the downstream task. In this article, we also examine the differences between the original \ac{MLM} task and the \ac{MLM} task with \ac{WWM}.

\section{Fine-tuning Techniques for NER}\label{sec:Fine-tuning}

The task of \ac{NER} is to detect entities, such as persons or places, which possibly consist of several words within a text.
As proposed in \citep{devlin2018bert}, the traditional approach for fine-tuning a \ac{BERT} to a classification task like \ac{NER} is to attach an additional feed-forward layer to a pre-trained \ac{BERT} which predicts token-wise labels.
In order to preserve and obtain information about the grouping of tokens into entities, \ac{IOB} tagging \citep{ramshaw_text_1999} is usually applied.
 \ac{IOB} tagging introduces two versions of each entity class, one marking the beginning of the entity and one representing the interior of an entity, and an ``other'' class, which all together results in a total of $\gamma=2e + 1$ tag classes where $e$ is the number of entity classes.
Table~\ref{tab:BItagscheme} shows an example in which the beginning token of an entity is prefixed with a ``B-'' and all other tokens with an ``I-''.

\begin{table}[ht]
\centering
\caption{
\ac{IOB} tagging example with unlabeled words (O) and  the two entities: ``location'' (Loc) and ``person'' (Per). The first tag of each entity is prefixed with ``B-'', while all following tokens of that entity are marked with an ``I-''.
The first row are the words of the sentence which are split into one or more tokens (second row).
The third row shows the tagged tokens based on the given entities (last row).
The example sentence can be translated as ``Peter lives in Frankfurt am Main''.
}
\begin{tabular}{l|c|c|c|cccc}
\toprule
    \textbf{Words} & Peter & lebt & in & \multicolumn{2}{c}{Frankfurt} & am & Main  \\
    \textbf{Tokens} & Peter & lebt & in & Frank & \_furt & am & Main \\
    \textbf{Tagged Tokens} & B-Per & O & O & B-Loc & I-Loc & I-Loc & I-Loc \\
    \textbf{Entities} & \multicolumn{1}{c|}{Person} &  &  &  \multicolumn{4}{c}{Location} \\
    \bottomrule
\end{tabular}
\label{tab:BItagscheme}
\end{table}

In compliance with the standard evaluation scheme of \ac{NER} tasks in \citep{sang2003introduction}, we compute an entity-wise $F_1$ score denoted by E-$F_1$.
Instead of computing a token- or word-wise $F_1$ score, E-$F_1$ evaluates a complete entity as true positive only if all tokens belonging to the entity are correct.
Our implementation of E-$F_1$ relies on the widely used Python library \emph{seqeval} \citep{seqeval}.

Usually, \ac{IOB} tagging is trained by a token-wise softmax cross-entropy loss.
However, this setup of one feed-forward layer and a cross-entropy loss does not take into account the context of the tokens forming an entity. In the following, we will call this default approach of fine-tuning the \ac{BERT} \acl{DFT}.
It can lead to inconsistent tagging, for example, an inner tag may only be preceded by an inner or beginning tag of the same entity, and thus results in a devastating impact on the E-$F_1$-score.
Therefore, we propose and compare three modified strategies that include context to prevent inconsistent \ac{NER} tagging during training or decoding.
The first approach is a modification of the \ac{IOB} tagging, the second proposal uses \acfp{LCRF}, the last attempt applies rules to fix a predicted tagging.

Most papers on \ac{BERT} models dealing with German \ac{NER}, for example \citep{chan2020german} or \citep{labusch_bert_2019}, do not focus on an investigation of different variants for fine-tuning.
However, there are already studies for \ac{NER} tasks in other languages \citep[e.g.][]{luoma_exploring_2020,souza_portuguese_2020} which show that the application of \acp{LCRF} can be beneficial for fine-tuning. \cite{souza_portuguese_2020} also investigated whether it is advantageous for the fine-tuning of \ac{BERT} models on NER tasks to link the pre-trained \ac{BERT} models with LSTM layers.
However, these experiments did not prove to be successful.


\subsection{Fine-tuning with \acs{CSE} tagging}\label{sec:ftcse}

In this section, we propose an alternative to the \ac{IOB} tagging which we call 
\acf{CSE} tagging.
The main idea is to split the task into three objectives as shown in Table~\ref{tab:example-cse}: finding start and end tokens, and learning the correct class.

\begin{table}[ht]
\label{cse-tagging}
\centering
\caption{
CSE tagging example.
Rows refer to the tokens and its respective target for Start, End, and Class.
}
\begin{tabular}{l|cccccccc}
\toprule
\textbf{Tokens} & Peter  & lebt & in & Frank & \_furt & am & Main  \\
\midrule
\textbf{Start} &  1  & 0 & 0 & 1 & 0 & 0 & 0 \\
\textbf{End} & 1  & 0 & 0 & 0 & 0 & 0 & 1 \\
\textbf{Class} & Per & O & O & Loc & Loc & Loc & Loc \\ 
\bottomrule
\end{tabular}
\label{tab:example-cse}
\end{table}

\ac{CSE} appends two additional dense layers with logistic-sigmoid activation to the last \ac{BERT} layer with scalar outputs, one for the start $p^{\text{start}}$, and one for the end $p^\text{end}$ token.
In summary, the complete output for an input sample consisting of $n$ tokens is $\left(\left(p^{\text{start}}_1, p^\text{end}_1, y_1\right), \left(p^{\text{start}}_2, p^\text{end}_2, y_2\right), \dots, \left(p^{\text{start}}_n, p^\text{end}_n, y_n\right) \right) \in \mathbb{R}^{n\times(2+e+1)}$ where $e+1$ is the number of possible entities and the ``other'' class.

The objective for $y_i$ is trained with softmax cross entropy as before but without the distinction between B- and I-, while the start and end vectors contribute extra losses $J^\text{start}$ and $J^\text{end}$:

\begin{equation}
    J^\text{start} = - \sum_{j=1}^{n}\bigl[
            t^{\text{start}}_j \cdot \log{(p^\text{start}_j)}
            + (1 - t^{\text{start}}_j) \cdot \log{(1 - p^\text{start}_j)}\bigr]\;,
\end{equation}
where $t^\text{start}$ and $p^\text{start}$ are the target and prediction vectors for start as shown in Table~\ref{tab:example-cse}. $J^\text{end}$ is defined analog.

Converting the \ac{CSE} into \ac{IOB} tagging is realized by accepting tokens which exceeds the threshold of 0.5 as start or end markers.
If an end marker is missing between two start markers, the position of the highest end probability between the two locations is used as an additional end marker.
This approach is applied analogue in reverse for missing start markers.
Finally, all class probabilities between each start and end marker pairs (including start and end) is averaged to obtain the entity class.
In conclusion, an inconsistent tagging is impossible.

\subsection{Fine-tuning with \acl{LCRF} with \ac{NER}-Rule ($\mathrm{LCRF}_\mathrm{NER}$)}\label{sec:ftlcrfner}

Another approach to tackle inconsistent \ac{IOB} tagging during fine-tuning of a \ac{BERT} is based on \acfp{LCRF} which are a modification of Conditional Random fields,  both proposed in \citep{lafferty_conditional_2001}.
\acp{LCRF} are a common approach to train neural networks that model a sequential task and are therefore well suited for fine-tuning \ac{NER}.
The basic idea is to take into account the classification of the neighboring sequence members when classifying an element of a sequence.

The output $Y=(y_1,y_2,\ldots,y_n)\in \mathbb{R}^{n\times \gamma}$ of our neural network for the \ac{NER} task consists of a sequence of $n$ vectors whose dimension corresponds to the number of classes $\gamma\in\mathbb{N}$. \ac{LCRF} introduce so-called transition values $\mathfrak{T}$ which are a matrix $W^{\mathfrak{T}}$ of trainable weights, in the basic approach:  $\mathfrak{T}:=W^{\mathfrak{T}}\in\mathbb{R}^{\gamma \times \gamma}$.
An entry $\mathfrak{T}_{i,j}$ of this matrix $\mathfrak{T}$ can be seen as the potential that a tag of class $i$ is followed by a tag of class $j$.
In one of the easiest forms of \acp{LCRF} which we choose, decoding aims to find the sequence $C^{p}:=\left\{c^{p}_1,c^{p}_2,\ldots,c^{p}_n \right\} \in \left\{1,2,\ldots,\gamma \right\}^{n}$ with the highest sum of corresponding transition values and elements of the corresponding output vectors as shown in eq. ~\eqref{eq:lcrf}.

\begin{equation}
\label{eq:lcrf}
 C^{p}:=\argmax_{C \in \{ 1, \ldots, \gamma \}^n}
 \left( 
 \sum\limits_{j=1}^{n}
 y_{j,c_j} 
 + 
 \sum\limits_{j=1}^{n-1}\mathfrak{T}_{c_j,c_{j+1}} 
 \right)
\end{equation}
Eq.~\eqref{eq:lcrf} is efficiently solved by the Viterbi-Algorithm  \citep[see e.g.][]{sutton_introduction_nodate}.
During training, a log-likelihood loss is calculated that takes into account the transition values $\mathfrak{T}$ and the network output $Y$.
\cite{sutton_introduction_nodate} provides a detailed description for its implementation. 

Since the \ac{IOB} tagging does not allow all possible transitions, \cite{lester-etal-2020-constrained} tried to simply ban these forbidden transitions completely by assigning fixed non-trainable high negative values to the associated entries in $\mathfrak{T}$. However, this did not lead to any improvement in performance, but they were able to show that this allows finetuning to converge faster when switching from the classic \ac{IOB} tagging to the more detailed IOBES tagging scheme~\citep{lester-etal-2020-constrained}.  In contrast to them, we extend the original \ac{LCRF} approach by explicitly modeling these forbidden transitions by adding additional trainable weights to the model when computing the transition values $\mathfrak{T}$.
In the following, we call our adapted algorithm $\mathrm{LCRF}_\mathrm{NER}$.

Assume an \ac{NER} task comprises the set of entities $\text{X}_1,\text{X}_2,\ldots,\text{X}_e$ which results in $\gamma=2e+1$ classes following the \ac{IOB} tagging scheme. 
Thus, beside a label $O$ for unlabeled elements, for each entity $\text{X}_i$ there is a begin label $\text{B-X}_i$ and an inner label $\text{I-}X_i$.
For simplicity, we order these classes by $\text{B-X}_1$,$\ldots$,$\text{B-X}_e$,$\text{I-X}_1$,$\ldots$,$\text{I-X}_e$,$O$, that is:

\begin{equation*}
 \text{Class $i$ belongs to label} \begin{cases}
\text{B-X}_i & \text{if $i\leq e$} \\
\text{I-X}_{i-e} & \text{if $e<i \leq 2e$} \\
O & \text{otherwise.} 
\end{cases}
\end{equation*}
With respect to this ordering, we introduce the matrix  $\mathfrak{F}\in\left\{0,1\right\}^{\gamma \times \gamma}$ of all forbidden transitions as

\begin{equation*}
    \mathfrak{F}_{i,j}= \begin{cases}
    1 & \text{if $e<j\leq 2e$ and $i\neq j$ and $i\neq j-e$} \\ 
    0 & \text{otherwise.}
    \end{cases}
\end{equation*}
Thus, an element $\mathfrak{F}_{i,j}$ is $1$, if and only if a tag of class $j$ can not follow on a tag of class $i$ in the given \ac{NER} task.
This maps the constraint that the label of the predecessor of an interior tag of label $\text{I-X}$ can only be the same interior label $\text{I-X}$ or the corresponding begin label $\text{B-X}$.

    \begin{figure}
        \begin{center}
            \includegraphics[width=0.9\textwidth]{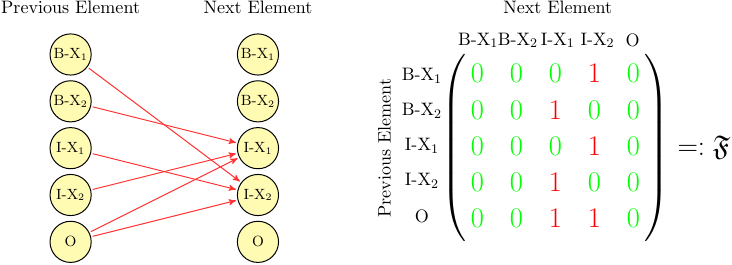}   

        \end{center}

        \caption{Example for the definition of the matrix $\mathfrak{F}$ of all forbidden transitions for two entities $\text{X}_1$, $\text{X}_2$. If we follow the IOB tagging scheme, red arrows mark forbidden transitions between two sequence elements that lead to an entry $1$ in $\mathfrak{F}$.}
        \label{fig:lcrf_ner}
    \end{figure}

In Figure \ref{fig:lcrf_ner} we illustrate the definition of $\mathfrak{F}$.

Likewise, we define the matrix $\mathfrak{A}\in\left\{0,1\right\}^{\gamma \times \gamma}$ by $\mathfrak{A}_{i,j}=1-\mathfrak{F}_{i,j}$ as the matrix of all allowed tag transitions.
$\mathrm{LCRF}_\mathrm{NER}$ introduces two additional trainable weights $\omega^{\mathfrak{F}}_\mathrm{factor},\omega^{\mathfrak{F}}_\mathrm{absolute} \in \mathbb{R}$ besides the weights $W^{\mathfrak{T}}$ and constructs $\mathfrak{T}$ by

\begin{equation}\label{eq:lcrfnertransvalue}
    \mathfrak{T}:=(\mathfrak{A}+\omega^{\mathfrak{F}}_\mathrm{factor}\mathfrak{F})\odot W^{\mathfrak{T}}-\omega^{\mathfrak{F}}_\mathrm{absolute}\mathfrak{F},
\end{equation}
where $\odot$ is the point-wise product.
If setting $\omega^{\mathfrak{F}}_\mathrm{factor}=1$ and $\omega^{\mathfrak{F}}_\mathrm{absolute}=0$ this defaults to the original \ac{LCRF} approach.
In this way, the model can learn an absolute penalty by $\omega^\mathfrak{F}_\mathrm{absolute}$ and a relative penalty by $\omega^\mathfrak{F}_\mathrm{factor}$ for forbidden transitions.
Note, that $\mathrm{LCRF}_\mathrm{NER}$ is mathematically equivalent to \ac{LCRF}, the only purpose is to simplify and to stabilize the training.

\subsection{Decoding with Entity-Fix Rule}\label{sec:ftfr}

Finally, we propose a rule-based approach to resolve inconsistent \ac{IOB} tagging which can for example occur if an I-X tag is subsequent to a token that is not I-X or B-X (for any possible entity class X).
Our so-called Entity-Fix rule replaces forbidden I-X tags with the tag of the previous token.
If the previous token has a B-X tag, the inserted token is converted to the corresponding I-X tag.
In the special case where an I-X tag is predicted at the start of the sequence, it is converted to B-X of the same class.
See Table~\ref{tab:example_fix_rule} for an example.
The advantage of this approach is that it can be applied as a post-processing step independent of training.
Furthermore, since only tokens which already form an incorrect entity are affected by this rule, the E-$F_1$ score can never decrease by applying it.
Note that this does not necessarily hold for the token-wise $F_1$ score, though.

    \begin{table}[ht]
        \caption{Example for Entity-Fix rule. Rows refer to the tokens, its respective target, prediction, and the prediction resulting from decoding with Entity-Fix rule. Changes are emphasized in bold.}
        \label{tab:example_fix_rule}
        \centering
        \begin{tabular}{l|ccccccc}
        \toprule
            \textbf{Tokens}     & Peter          & lebt & in & Frank & \_furt & am & Main  \\ \midrule
            \textbf{Target}     & B-Per          & O   & O  & B-Loc & I-Loc & I-Loc & I-Loc \\
            \textbf{Prediction}& \textbf{I-Per} & O   & O  & B-Loc & \textbf{I-Org} &\textbf{I-Org} & I-Loc\\
            \textbf{Prediction with 
            Fix-Rule}   & \textbf{B-Per} & O   & O  & B-Loc & \textbf{I-Loc} & \textbf{I-Loc} & I-Loc \\
            \bottomrule
        \end{tabular}
    \end{table}

\section{\ac{BERT} Architecture with  \acf{WWA}}\label{sec:WWA}
In this section, we describe our proposed word-wise attention layers used by some of our \ac{BERT} models during pre-training and fine-tuning. This \acf{WWA} was inspired by the benefits of the \acf{WWM}. It comprises two components: the first one called $\mathrm{mha}_{\mathrm{wwa}}$ applies traditional multi-head attention on words instead of tokens, while the second component is a windowed attention module called $\mathrm{mha}_{\mathrm{wind}}$.

\paragraph{Traditional Approach}
In opposite to current \ac{NLP} network architectures, previous approaches for tokenizing text  \citep[e.g.][]{mikolov2013distributed} did not apply a tokenizer to break down each word of a sentence into possibly more than one token. Instead, they trained representations for a fixed vocabulary of words. 
The major drawback was that this required a large vocabulary and out-of-vocabulary words could not be represented.
Modern approaches tokenize words by a vocabulary of subwords which allows to compose unknown words by known tokens.
However, when combined with Transformers, attention is computed between pairs of tokens. As a consequence the number of energy values (see eq.~\eqref{eq:attenergy}) to be calculated increases quadratically with sequence length resulting in a large increase of memory and computation time for long sequences.

There exist different approaches to tackle this problem.
The most prominent ones are BigBird~\citep{zaheer2020big} and Longformer~\citep{beltagy2020longformer}. In their work, the focus is on pure sparse attention strategies: Instead of a full attention, they try to omit as many calculations of energy values as possible, so that as little performance as possible is lost.
Instead, we propose to rejoin tokens into word-based tokens which also has a quadratic dependence on the sequence length but by a lower slope.

\paragraph{Our Methodology}
The purpose of the first module, $\mathrm{mha}_{\mathrm{wwa}}$, is to map tokens back to words and then to compute a word-wise attention.
However, since $\mathrm{mha}_{\mathrm{wwa}}$ loses information about the order of tokens within a word, we introduce $\mathrm{mha}_{\mathrm{wind}}$ as additional component which acts on the original tokens.
$\mathrm{mha}_{\mathrm{wind}}$ scales linearly with the sequence length since only a window of tokens is taken into account when computing the energy vectors.
In summary, $\mathrm{mha}_{\mathrm{wwa}}$ learns the global coarser dependence of words whereas $\mathrm{mha}_{\mathrm{wind}}$ allows to resolve and learn relations of tokens but only in a limited range.
In the following, we first describe $\mathrm{mha}_{\mathrm{wwa}}$ and then $\mathrm{mha}_{\mathrm{wind}}$.

Let $T$ denote the input of our \ac{BERT} model which is a part of text and can thus be seen as a sequence of words $T=(w_1,w_2,\ldots, w_m)$ with $m\in\mathbb{N}$.
Similar to a classical \ac{BERT}, a tokenizer $\mathcal{T}$ transforms $T$ into a sequence of tokens $\mathcal{T}(T)=:t=(t_1,t_2,\ldots,t_n) \in \mathbb{N}^n$ with $m\leq n$ because we only consider traditional tokenizers that encode the text word-wisely by decomposing a word into one or more tokens.
Such a tokenizer provides a mapping function $F_{T,\mathcal{T}}: \left\{1,2,\ldots,n\right\} \rightarrow \left\{1,2,\ldots,m\right\} $ which uniquely maps an index $i$ of the token sequence $t$ to the index $j$ of its respective word $w_j$. 

\begin{figure}
\begin{center}
    \resizebox{\textwidth}{!}{
\begin{tikzpicture}[scale=1,>=stealth',semithick,auto]

    \tikzstyle{obj}  = [rounded rectangle, minimum width=3cm, minimum height = 0.6cm, draw, inner sep=0pt, fill =gray!50]
    \tikzstyle{rec}  = [rounded rectangle, minimum width=6.7cm, minimum height = 0.8cm, draw, inner sep=0pt, fill = green!30]
    \tikzstyle{tokseq}  = [rounded rectangle, rounded corners=5pt, minimum width=1.25cm, minimum height = 0.5cm, draw, inner sep=0pt, fill = orange!30]
    \tikzstyle{wordseq}  = [rounded rectangle, rounded corners=4pt, minimum width=3cm, minimum height = 0.5cm, draw, inner sep=0pt, fill = red!30]

    \node (tradtitle) at (1.625,8.5) {\Large Classical Multi-Head Attention};
    \node[tokseq] (x1t) at (0,-0.5) {$x_1$};
    \node[tokseq] (x2t) at (1.25,-0.5) {$x_2$};
    \node[tokseq] (xw1dotst) at (2.5,-0.5) {$\ldots$};
    \node[tokseq] (xnt) at (3.75,-0.5) {$x_n$};
    \node[rec] (mhat) at (1.625,4) {\large Multi-Head Attention Layer ($\mathrm{mha}$)};
    \node[tokseq] (y1t) at (0,7.5) {$y_1$};
    \node[tokseq] (y2t) at (1.25,7.5) {$y_2$};
    \node[tokseq] (yw1dotst) at (2.5,7.5) {$\ldots$};
    \node[tokseq] (ynt) at (3.75,7.5) {$y_n$};
    \path[<-] (mhat) edge (x1t) edge (x2t) edge (xnt);
    \path[->] (mhat) edge (y1t) edge (y2t) edge (ynt);

    \node (tradtitle) at (15,8.5) {\Large \acl{WWA}};

    \node[tokseq] (x1o) at (7.5,-0.5) {$x_1$};
    \node[tokseq] (x2o) at (8.75,-0.5) {$x_2$};
    \node[tokseq] (xw1dotso) at (10,-0.5) {$\ldots$};
    \node[tokseq] (xf1o) at (11.25,-0.5) {$x_{F_1}$};
    \node[tokseq] (xf11o) at (12.5,-0.5) {$x_{F_1+1}$};
    \node[tokseq] (xf12o) at (13.75,-0.5) {$x_{F_1+2}$};
    \node[tokseq] (xw2dotso) at (15,-0.5) {$\ldots$};
    \node[tokseq] (xf1f2o) at (16.25,-0.5) {$x_{F_1+F_2}$};
    \node[tokseq] (xwdotso) at (17.5,-0.5) {$\ldots$};
    \node[tokseq] (xe1o) at (18.75,-0.5) {$\ldots$};
    \node[tokseq] (xe2o) at (20,-0.5) {$\ldots$};
    \node[tokseq] (xwmdotso) at (21.25,-0.5) {$\ldots$};
    \node[tokseq] (xno) at (22.5,-0.5) {$x_n$};
    \node[wordseq] (xh1) at (10.5,2) {$\hat{x}_1$};
    \node[wordseq] (xh2) at (13.5,2) {$\hat{x}_2$};
    \node[wordseq] (xhwdots) at (16.5,2) {$\ldots$};
    \node[wordseq] (xhm) at (19.5,2) {$\hat{x}_m$};
    \node[rec] (mhao) at (15,4) {\large Multi-Head Attention Layer ($\mathrm{mha}$)};
    \node[wordseq] (yh1) at (10.5,6) {$\hat{y}_1$};
    \node[wordseq] (yh2) at (13.5,6) {$\hat{y}_2$};
    \node[wordseq] (yhwdots) at (16.5,6) {$\ldots$};
    \node[wordseq] (yhm) at (19.5,6) {$\hat{y}_m$};
    \node[tokseq] (z1o) at (7.5,7.5) {$z_1$};
    \node[tokseq] (z2o) at (8.75,7.5) {$z_2$};
    \node[tokseq] (zw1dotso) at (10,7.5) {$\ldots$};
    \node[tokseq] (zf1o) at (11.25,7.5) {$z_{F_1}$};
    \node[tokseq] (zf11o) at (12.5,7.5) {$z_{F_1+1}$};
    \node[tokseq] (zf12o) at (13.75,7.5) {$z_{F_1+2}$};
    \node[tokseq] (zw2dotso) at (15,7.5) {$\ldots$};
    \node[tokseq] (zf1f2o) at (16.25,7.5) {$z_{F_1+F_2}$};
    \node[tokseq] (zwdotso) at (17.5,7.5) {$\ldots$};
    \node[tokseq] (ze1o) at (18.75,7.5) {$\ldots$};
    \node[tokseq] (ze2o) at (20,7.5) {$\ldots$};
    \node[tokseq] (zwmdotso) at (21.25,7.5) {$\ldots$};
    \node[tokseq] (zno) at (22.5,7.5) {$z_n$};

    \path[<-] (xh1) edge (x1o) edge (x2o) edge (xf1o);
    \path[<-] (xh2) edge (xf11o) edge (xf12o) edge (xf1f2o);
    \path[<-] (xhm) edge (xe1o) edge (xe2o) edge (xno);
    \path[<-] (mhao) edge (xh1) edge (xh2) edge (xhm);

    \path[->] (yh1) edge (z1o) edge (z2o) edge (zf1o);
    \path[->] (yh2) edge (zf11o) edge (zf12o) edge (zf1f2o);
    \path[->] (yhm) edge (ze1o) edge (ze2o) edge (zno);
    \path[->] (mhao) edge (yh1) edge (yh2) edge (yhm);

    \node[obj] (avg1) at (10,0.8) {Average};
    \node[obj] (avg2) at (13.8,0.8) {Average};
    \node[obj] (avgm) at (20,0.8) {Average};

\end{tikzpicture}}
\end{center}
 \caption{Left: Multi-Head Attention. Right: Our concept of \acl{WWA}, where classical Multi-Head Attention is applied on words instead of tokens. Orange: Members of sequences, whose length is the number of tokens $n$. Red: Members of sequences, whose length is the number of words $m$. For a better overview we define $F_j:=\left|\left\{i:F_{T,\mathcal{T}}(i)=j\right\}\right|$ as the number of tokens the $j$'th word consists of.
 }\label{fig:wwa}
\end{figure}
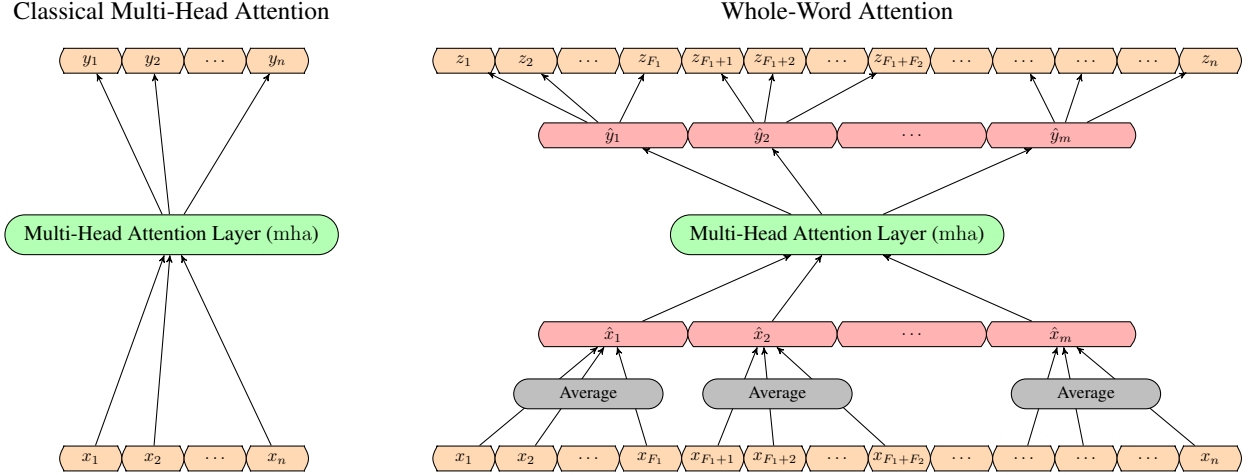

Each encoder layer $\ell$ of the classical \ac{BERT} architecture contains a multi-head attention layer $\mathrm{mha}^{\ell}$ which maps its input sequence $X_{T,\mathcal{T}}^{\ell}=(x_1^{\ell},x_2^{\ell},\ldots,x_n^{\ell})\in \mathbb{R}^{n\times d}$ to an output $Y_{T,\mathcal{T}}^{\ell}$ of equal length $n$ and dimension $d$:

\begin{equation*}
\mathrm{mha}^{\ell}(X_{T,\mathcal{T}}^{\ell})=Y_{T,\mathcal{T}}^{\ell}=(y_1^{\ell},y_2^{\ell},\ldots,y_n^{\ell})\in \mathbb{R}^{n\times d},
\end{equation*}
where the $i$th output vector $y^{\ell}_i$ is defined as the concatenation of the resulting vectors for every attention head computed by equation~\eqref{eq:RelAtt}.
Our $\mathrm{mha}_{\mathrm{wwa}}$ layer modifies this by applying attention only on the sequence $\hat{X}_{T,\mathcal{T}}^{\ell}=(\hat{x}^{\ell}_1,\hat{x}^{\ell}_2,\ldots,\hat{x}^{\ell}_m)\in \mathbb{R}^{m\times d}$, where 

\begin{equation}\label{eq:wwavg}
\hat{x}_j:=\frac{1}{\left|\left\{i:F_{T,\mathcal{T}}(i)=j\right\}\right|}\sum\limits_{i: F_{T,\mathcal{T}}(i)=j}x_i
\end{equation} 
and $\left\{i: F_{T,\mathcal{T}}(i)=j\right\}$ is the set of all tokens $i$ belonging to the word $j$. 
In other words, we average the corresponding token input vectors for each word.
Next, we apply $\mathrm{mha}$ on $\hat{X}_{T,\mathcal{T}}^{\ell}$ yielding the output

\begin{equation*}
\mathrm{mha}^{\ell}(\hat{X}_{T,\mathcal{T}}^{\ell})=:\hat{Y}_{T,\mathcal{T}}^{\ell}=(\hat{y}_1^{\ell},\hat{y}_2^{\ell},\ldots,\hat{y}_m^{\ell})\in \mathbb{R}^{m\times d}
\end{equation*}
which is a sequence of length $m$ only.
Finally, to again obtain a sequence of length $n$, we transform the output sequence back to the length $n$ by repeating the output vector for each word according to the number of associated tokens.
Thus, the final output of a layer $\mathrm{mha}_{\mathrm{wwa}}^\ell$ is defined as

\begin{equation*}
   \mathrm{mha}_{\mathrm{wwa}}^\ell (X_{T,\mathcal{T}}^{\ell}):=Z_{T,\mathcal{T}}^{\ell}=(z_1^{\ell},z_2^{\ell},\ldots,z_n^{\ell})\in \mathbb{R}^{n\times d}
\end{equation*} 
where $z_i^{\ell}:=\hat{y}_{F_{T,\mathcal{T}}(i)}^{\ell}$. 
See Figure \ref{fig:wwa} for an illustration of the concept described above.

We perform positional encoding by utilizing relative positional encoding because absolute positional encoding adds the positional vectors to the token sequence directly after the embedding.
This is not compatible with \ac{WWA} of eq.~\eqref{eq:wwavg}. Instead, 
relative positional encoding adds a word-wise relative positional encoding to the vectors of the word-wise sequence within $\mathrm{mha}^{\ell}(\hat{X}_{T,\mathcal{T}}^{\ell})$.

As an experiment, we also pre-trained a BERT which solely uses  $\mathrm{mha}_{\mathrm{wwa}}$ layers for attention. However, it was already apparent in the pre-training that it could only achieve a very low \acf{MLM} accuracy. The main reason for this is that $Z_{T,\mathcal{T}}^{\ell}$ does not take into account any information about the position of the tokens within a word since the output elements of them are equal which is why these tokens are no longer related to each other via attention.
To tackle this problem, for each $\ell$, we introduce a second multi-head attention layer $\mathrm{mha}_{\mathrm{wind}}^\ell$ based on windowed attention as used in \citep{beltagy2020longformer,zaheer2020big}.
Because the sole purpose of $\mathrm{mha}_{\mathrm{wind}}^\ell$ is to map the relationships and positions of the tokens within a word in the model, we use a very small sliding window size of $\omega=5$ tokens in each direction.
Hence, in contrast to \cite{beltagy2020longformer,zaheer2020big}, we also do not arrange our input sequence into blocks or chunks.

Formally, we define $\mathrm{mha}_{\mathrm{wind}}$ as

\begin{equation*}
   \mathrm{mha}_{\mathrm{wind}}^\ell (X_{T,\mathcal{T}}^{\ell}):=Y_{T,\mathcal{T}}^{\ell}=(y_1^{\ell},y_2^{\ell},\ldots,y_n^{\ell})\in \mathbb{R}^{n\times d}
\end{equation*}
where the $i$th output vector $y^{\ell}_i$ is the concatenation of the resulting vectors $y^{\ell,h}_i$ for every attention head $h$.
Each $y^{\ell,h}_i$ is given by

\begin{equation*}
    y^{\ell,h}_i=\sum\limits_{j=i-\omega}^{i+\omega}\alpha_{i,j}^h\left(W^{V,h} x_{j}+d_{j-i}^{V,h}\right)
\end{equation*}
where $\alpha_{i,j}^h=\frac{exp\left(e_{i,j}^h\right)}{\sum\limits_{k=1}^nexp\left(e_{i,k}^h\right)}$,$W^{V,h}\in\mathbb{R}^{d_{model}\times d_v}$. 
The energy values $e_{i,j}^h$ for each attention head $h$ are defined as in equation~\eqref{eq:attenergy} and the distance vectors $d_{j-i}^{V,h}, d_{j-i}^{K,h}$ are given by the equations~\eqref{eqn:distvectK} and~\eqref{eqn:distvectV}.

In summary, in our \ac{BERT} model using \ac{WWA}, the total output of the $\ell$-th attention layer of the $\ell$-th encoder layer is, after adding the input sequence $X_{T,\mathcal{T}}^{\ell}$ as residual,

\begin{equation*}
    X_{T,\mathcal{T}}^{\ell} + \mathrm{mha}_{\mathrm{wwa}}^\ell(X_{T,\mathcal{T}}^{\ell}) + \mathrm{mha}_{\mathrm{wind}}^\ell(X_{T,\mathcal{T}}^{\ell})
\end{equation*}
compared to the original approach

\begin{equation*}
    X_{T,\mathcal{T}}^{\ell} + \mathrm{mha}^\ell(X_{T,\mathcal{T}}^{\ell}).
\end{equation*}
$\mathrm{mha}_{\mathrm{wind}}$ introduces additional trainable variables compared to the traditional Transformer architecture.
However, the number of energy values increases only linearly compared to the sequence length with respect to the window size $\omega$ by a factor $2\omega+1$. Hence, the impact on the memory can be neglected for long sequences and small $\omega$.
In order to quantify the overall reduction of memory consumption using \ac{WWA}, we provide an example using our tokenizer\footnote{We use a tokenizer with a vocabulary of about 30,000 as common in many \ac{BERT} models.} built on the German Wikipedia which transforms on average a sequence of $m$ words in $n\approx1.5 m = \frac{3}{2}m$ tokens.
Therefore, while the traditional multi-head attention layer calculates $n^2$ energy values per each head, our \ac{WWA} approach only requires 

\begin{equation}\label{eq:wordtokencomp}
    \overbrace{m^2}^{\mathrm{mha}_{\mathrm{wwa}}}+\overbrace{(2\omega+1)n}^{\mathrm{mha}_{\mathrm{wind}}}\approx\frac{4}{9}n^2+(2\omega+1)n
\end{equation}
energy values.
Therefore, for large $n$ (and small $\omega$), the number of energy values to be calculated is more than halved.

\section{Experiments}\label{sec:Experiments}

To evaluate our proposed methods for German \ac{NER} tasks, we conducted several experiments.
First, we compared the pre-training variants presented in Section \ref{sec:Pre-training} and then fine-tuning techniques presented in Section \ref{sec:Fine-tuning}.
Afterwards, we applied \acf{WWA} on the overall training of the \ac{NER} task.
Finally, we discuss our results in relation to the state-of-the-art models of the \ac{LER} and GermEval tasks.

\subsection{Comparing Pre-training Techniques}\label{sec:exppre-train}

In our first experiments, we investigated which of the known pre-training techniques (see Section~\ref{sec:Pre-training}) yields the best models for a subsequent fine-tuning on German \ac{NER} tasks.
For this purpose, we combined the three presented pre-training tasks (\acf{MLM}, \ac{MLM} with \acf{SOP}, and \ac{MLM} with \acf{NSP}) with relative and absolute positional encoding and optionally enabled \acf{WWM}.
For each resulting combination, we pre-trained a (small) \ac{BERT} with a hidden size of 512 with 8 attention heads on 6 layers for 500 epochs and 100,000 samples per epoch.
Pre-training was performed with a batch size of 48 and a maximal sequence length of 320 tokens which is limited by the 11\,GB memory of one GPU (RTX 2080 Ti). 
Each of the 12 resulting \acp{BERT} was then fine-tuned using the \acl{DFT} approach (described in Section \ref{sec:Fine-tuning}) on the five German \ac{NER} tasks described in Section~\ref{sec:NERtasks}.
For fine-tuning, we chose a batch size of 16 and trained by default for 30 epochs and 5,000 samples per epoch.
The number of epochs was increased to 50 for the larger \ac{LER} tasks.
Each fine-tuning run was performed three times and its average result was reported in Table~\ref{tab:exppre-train} together with its standard deviation $\sigma$.

\begin{table}[tbp]
\caption{Columns refer to the average E-$F_1$ score \protect \citep[cf.][]{sang2003introduction} of three fine-tuning runs with \acl{DFT} (see section \ref{sec:Fine-tuning}) for five datasets and its standard deviation $\sigma$ multiplied by $100$. Rows refer to the respective pre-training task, absolute or relative positional encoding (PE), and use of \acf{WWM}; best results within $2\cdot\sigma$ of the maximum (best) are emphasized}
\label{tab:exppre-train}
\centering
\resizebox{\textwidth}{!}{
\begin{tabular}{p{1.8cm}p{0.5cm}c|rr|rr|rr|rr|rr|r}
\toprule
\textbf{Pre-train task} & \textbf{PE} & \textbf{WWM} & \multicolumn{2}{l|}{\textbf{GermEval}} & \multicolumn{2}{l|}{\textbf{H.Arendt}}  & \multicolumn{2}{l|}{\textbf{Sturm}} & \multicolumn{2}{l|}{\textbf{LER CG}} & \multicolumn{2}{l|}{\textbf{LER FG}} & \multicolumn{1}{l}{\textbf{Average}} \\ 
 &  &  & \multicolumn{1}{l}{E-$F_1$} & \multicolumn{1}{l|}{$100\sigma$} & \multicolumn{1}{l}{E-$F_1$} & \multicolumn{1}{l|}{$100\sigma$} & \multicolumn{1}{l}{E-$F_1$} & \multicolumn{1}{l|}{$100\sigma$} & \multicolumn{1}{l}{E-$F_1$} & \multicolumn{1}{l|}{$100\sigma$} & \multicolumn{1}{l}{E-$F_1$} & \multicolumn{1}{l|}{$100\sigma$} & \multicolumn{1}{l}{E-$F_1$} \\ \midrule
MLM & abs. & - & 0.7785 & 0.09 & 0.7600 & 1.28 & 0.8236 & 1.20 & 0.9061 & 0.48 & 0.8969 & 0.37 & 0.8330 \\ 
MLM & abs. & \checkmark & 0.7901 & 0.21 & 0.7681 & 0.40 & 0.8102 & 1.73 & 0.9192 & 0.12 & 0.9028 & 0.21 & 0.8381 \\ 
MLM & rel. & - & 0.7852 & 0.16 & 0.7674 & 0.54 & 0.8011 & 1.45 & 0.9198 & 0.34 & 0.9144 & 0.27 & 0.8376 \\ 
MLM & rel. & \checkmark & \textbf{0.8086} & 0.24 & \textbf{0.7741} & 0.79 & \textbf{0.8555} & 0.34 & \textbf{0.9347} & 0.30 & \textbf{0.9206} & 0.14 & \textbf{0.8587} \\ \midrule
MLM, NSP & abs. & - & 0.7609 & 0.10 & 0.7688 & 1.21 & 0.8085 & 0.20 & 0.9067 & 0.36 & 0.8928 & 0.56 & 0.8275 \\ 
MLM, NSP & abs. & \checkmark & 0.7484 & 1.52 & 0.7621 & 0.40 & 0.8110 & 1.38 & 0.9047 & 0.30 & 0.8930 & 0.30 & 0.8238 \\ 
MLM, NSP & rel. & - & 0.7802 & 0.40 & 0.7471 & 0.48 & 0.8172 & 1.37 & 0.9193 & 0.39 & 0.9105 & 0.18 & 0.8348 \\ 
MLM, NSP & rel. & \checkmark & 0.7750 & 0.23 & 0.7564 & 1.15 & 0.8191 & 0.40 & 0.9168 & 0.07 & 0.9044 & 0.38 & 0.8343 \\ \midrule
MLM, SOP & abs. & - & 0.7530 & 0.18 & 0.7669 & 0.71 & 0.7942 & 1.20 & 0.8979 & 0.25 & 0.8817 & 0.14 & 0.8188 \\ 
MLM, SOP & abs. & \checkmark & 0.7355 & 0.38 & 0.7590 & 0.21 & 0.7995 & 2.40 & 0.9047 & 0.17 & 0.8950 & 0.35 & 0.8187 \\ 
MLM, SOP & rel. & - & 0.7745 & 0.58 & 0.7548 & 0.52 & 0.8052 & 0.62 & 0.9176 & 0.24 & 0.9040 & 0.48 & 0.8312 \\ 
MLM, SOP & rel. & \checkmark & 0.7863 & 0.31 & \textbf{0.7807} & 0.41 & \textbf{0.8550} & 0.15 & 0.9246 & 0.24 & 0.9122 & 0.38 & 0.8518 \\ 
\bottomrule
\end{tabular}}
\end{table}

The first thing we can see in Table~\ref{tab:exppre-train} is that, as expected, the standard deviation for the \ac{NER} tasks with a smaller ground truth (mainly Sturm but also H.Arendt) is higher than for \ac{NER} tasks with larger ground truth (GermEval, both \ac{LER} variants). The fluctuations of the different fine-tuning runs are nevertheless within a reasonable range.

The results in Table~\ref{tab:exppre-train} reveal that the best performing \acp{BERT} for \ac{NER} tasks used relative positional encoding, \ac{WWM}, and were solely pre-trained with \ac{MLM}.
This is plausible since the samples, the academical \ac{NER} datasets consists of, are only single sentences and hence using additional sentence-spanning losses during pre-training (\ac{SOP} or \ac{NSP}) is even harmful for these downstream tasks.

\begin{figure}[t]
\includegraphics[width=0.99 \textwidth]{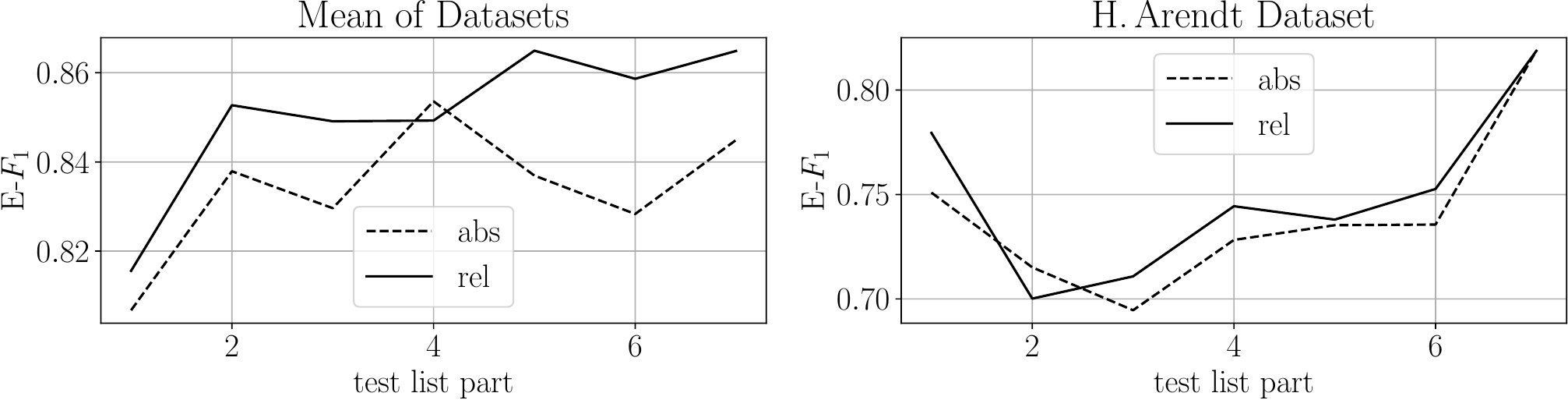}
\caption{E-$F_1$ score for a \ac{BERT} trained on \acf{MLM} task with \acf{WWM} for relative and absolute positional encoding fine-tuned with \acl{DFT} (see section \ref{sec:Fine-tuning}) tested on seven parts of the test list, which are sorted by token length, starting with short sequences. Left shows the mean over all datasets, Right shows the results on one dataset (H.\,Arendt).}
\label{fig:error_seqlen}
\end{figure}

Furthermore, our experiments show that relative positional encoding yields significantly better results than absolute positional encoding.
This is an interesting observation since the experiments from \cite{rosendahl2019analysis} stated that relative positional encoding only performs better when applied to sequences with length longer than those on which the network was previously trained.
To investigate this in detail, we performed an analysis of the E-$F_1$ score in dependence of the sequence length.
We sorted the samples in the test list for each dataset by token length in increasing order and then split them into seven parts of equal number of samples.
The left chart of Figure~\ref{fig:error_seqlen} shows the averaged E-$F_1$ score of the five datasets for each part.
We observe that relative encoding (rel) is outperforming absolute encoding (abs) for almost any sequence length.
For long sequences (last part), the gap between the two approaches increases which is expected.
The right chart of Figure~\ref{fig:error_seqlen} shows the E-$F_1$ score for each part of only the H.\,Arendt dataset.
For this dataset, the discrepancy between relative and positional encoding is very small which shows that the benefit of relative positional encoding highly depends on the dataset.
Nevertheless, on average, our experiments suggest to use relative positional as the mean of choice for German \ac{NER} tasks.

Furthermore, we observe that \ac{WWM} together with relative positional encoding led to significant improvements when combined with \ac{MLM} with or without \ac{SOP} as pre-training task.
In summary, our best setup combined solely \ac{MLM}, relative positional encoding, and \ac{WWM}.

\subsection{Comparing Fine-tuning Techniques}
In this section, we examine our four different variants of fine-tuning: \acl{DFT} (see section \ref{sec:Fine-tuning}), \acf{CSE} (see Section \ref{sec:ftcse}), and \acf{LCRF} or $\text{\ac{LCRF}}_\text{\ac{NER}}$ (see Section \ref{sec:ftlcrfner}). 
The new fine-tuning techniques \ac{CSE} and $\text{\ac{LCRF}}_\text{\ac{NER}}$ were specifically designed to help the network learn the structure of \ac{IOB} tagging.

First, we evaluated the impact of the fine-tuning method in dependence of the pre-training techniques examined in Section~\ref{sec:exppre-train}.
For this purpose, we fine-tuned all \ac{BERT} models in analogy to Section~\ref{sec:exppre-train} on all five \ac{NER} tasks using the four fine-tuning methods mentioned.
The individual results shown in Table~\ref{tab:expftavg} are averaged across the five tasks whereby the outcome of each task is the mean of three runs.

\begin{table}[tbp]
\caption{
Average E-$F_1$ score of all five \ac{NER} tasks and three fine-tuning runs. Rows refer to the respective pre-training task, absolute or relative positional encoding (PE), and use of \acf{WWM}; Columns refer to the fine-tuning methods \acf{DFT}, \acf{CSE}, \acf{LCRF} and $\text{LCRF}_\text{NER}$; best results per column are emphasized.}
\label{tab:expftavg}
\centering
\begin{tabular}{p{2cm}p{0.5cm}c|rrrr}
\toprule
\textbf{Pre-training task} & \textbf{PE} & \textbf{WWM} & \multicolumn{1}{c}{\textbf{Avg. \ac{DFT}}} & \multicolumn{1}{c}{\textbf{Avg. \ac{CSE}}} & \multicolumn{1}{c}{\textbf{Avg. \ac{LCRF}}} & \multicolumn{1}{c}{\textbf{Avg. $\text{LCRF}_\text{NER}$}} \\
\midrule
MLM & abs. & - & 0.8330 & 0.8608 & 0.8455 & 0.8469 \\ 
MLM & abs. & \checkmark & 0.8381 & 0.8635 & 0.8443 & 0.8516 \\ 
MLM & rel. & - & 0.8376 & 0.8682 & 0.8477 & 0.8539 \\ 
MLM & rel. & \checkmark & \textbf{0.8587} & \textbf{0.8785} & \textbf{0.8623} & \textbf{0.8651} \\ \midrule
MLM, NSP & abs. & - & 0.8275 & 0.8536 & 0.8346 & 0.8418 \\ 
MLM, NSP & abs. & \checkmark & 0.8238 & 0.8543 & 0.8358 & 0.8416 \\ 
MLM, NSP & rel. & - & 0.8348 & 0.8639 & 0.8457 & 0.8471 \\ 
MLM, NSP & rel. & \checkmark & 0.8343 & 0.8595 & 0.8417 & 0.8459 \\ \midrule
MLM, SOP & abs. & - & 0.8188 & 0.8515 & 0.8292 & 0.8325 \\ 
MLM, SOP & abs. & \checkmark & 0.8187 & 0.8597 & 0.8312 & 0.8369 \\
MLM, SOP & rel. & - & 0.8312 & 0.8598 & 0.8450 & 0.8473 \\ 
MLM, SOP & rel. & \checkmark & 0.8518 & 0.8714 & 0.8557 & 0.8605 \\ \midrule
\multicolumn{ 3}{l|}{Overall Average} & 0.8340 & 0.8608 & 0.8432 & 0.8476 \\ 
\bottomrule
\end{tabular}

\end{table}

Regardless of the choice of the fine-tuning method, the results confirm that pre-trained \acp{BERT} that were only pre-trained with \ac{MLM}, use relative Positional Encoding, and use \ac{WWM} are the most suitable for German \ac{NER} tasks.
Therefore, we will only consider this fine-tuning setup in the following.

Furthermore, the results show that $\text{LCRF}_\text{NER}$ consistently outperforms \ac{LCRF}. 
We think that the introduction of the new weights $\omega^{\mathfrak{F}}_\mathrm{factor},\omega^{\mathfrak{F}}_\mathrm{absolute}$ (see eq.~\ref{eq:lcrfnertransvalue}) enables the fine-tuning with $\text{LCRF}_\text{NER}$ to outperform \ac{LCRF} because the network can learn more easily to avoid inconsistent tagging.
However, \ac{CSE} performs best on average. 
We suspect that this is due to the fact that the way of decoding in \ac{CSE} completely prevents inconsistent tagging.

Table \ref{tab:expftfr} provides more details for the best setup by listing separate results per \ac{NER} task.
Furthermore, we examined the influence of applying the rule based Entity-Fix (see Section \ref{sec:ftfr}) as a post-processing step.

\begin{table}[htpb]
\caption{
Average E-$F_1$ score of three fine-tuning runs on \acp{BERT} pre-trained with \acf{MLM}, used relative positional encoding, and \acf{WWM}. Rows refer to the fine-tuning methods \acf{DFT}, \acf{CSE}, \acf{LCRF} and $\text{LCRF}_\text{NER}$; Columns refer to the \ac{NER} task; best results per column are emphasized.}
\label{tab:expftfr}
\centering
\resizebox{\textwidth}{!}{
\begin{tabular}{p{2cm}c|rrrrrr}
\toprule
\textbf{Fine-tuning task} & \multicolumn{1}{p{2cm}|}{\textbf{Entity-Fix rule}} & \multicolumn{1}{c}{\textbf{GermEval}} & \multicolumn{1}{c}{\textbf{H.\,Arendt}} & \multicolumn{1}{c}{\textbf{Sturm}} & \multicolumn{1}{c}{\textbf{\ac{LER} CG}} & \multicolumn{1}{c}{\textbf{\ac{LER} FG}} & \multicolumn{1}{c}{\textbf{Average}} \\ \midrule
\ac{DFT} & - & 0.8086 & 0.7741 & 0.8555 & 0.9347 & 0.9206 & 0.8587 \\ 
\ac{DFT} & \checkmark & 0.8408 & 0.7903 & 0.8706 & 0.9474 & 0.9427 & 0.8783 \\ \midrule
\ac{CSE} & - & 0.8397 & \textbf{0.8048} & 0.8647 & 0.9429 & 0.9401 & 0.8785 \\ 
\ac{CSE} & \checkmark & 0.8397 & \textbf{0.8048} & 0.8647 & 0.9429 & 0.9401 & 0.8785 \\ \midrule
\ac{LCRF} & - & 0.8216 & 0.7822 & 0.8453 & 0.9365 & 0.9261 & 0.8623 \\ 
\ac{LCRF} & \checkmark & 0.8422 & 0.7941 & 0.8629 & 0.9477 & 0.9410 & 0.8776 \\ \hline
$\text{LCRF}_\text{NER}$ & - & 0.8220 & 0.7857 & 0.8508 & 0.9394 & 0.9278 & 0.8651 \\ 
$\text{LCRF}_\text{NER}$ & \checkmark & \textbf{0.8448} & 0.7999 & \textbf{0.8783} & \textbf{0.9488} & \textbf{0.9455} & \textbf{0.8823} \\ \bottomrule

\end{tabular}}

\end{table}

The Entity-Fix rule was specifically designed to fix inconsistencies that otherwise would lead to an error on the metric. 
Since the \ac{CSE} decoding already includes the rules of the metric, no changes are present if applying the Entity-Fix rule.
This post-processing step led to clear improvements on all other fine-tuning methods.  
Surprisingly, although \ac{LCRF} and $\text{LCRF}_\text{NER}$ were specially designed to learn which consecutive tags were not allowed in the sequence, they still show significant improvements with the Entity-Fix rule. Thus, they were not able to learn the structure of the \ac{IOB} tagging scheme sufficiently.

The general result is that $\text{LCRF}_\text{NER}$ represents the best fine-tuning method if combined with the Entity-Fix rule.

Additionally, we examined why, in contrast to all other tasks, the H.\,Arendt task with \ac{CSE} yielded slightly better results than with $\text{LCRF}_\text{NER}$ and Entity-Fix rule by comparing the errors made on the test set.
Unfortunately, no explanation could be found in the data.
Nevertheless, $\text{LCRF}_\text{NER}$ emerges as the best fine-tuning method from our experiments in general.

\subsection{Results of \acf{WWA}}
Next, we investigated the influence of replacing the original multi-head attention layers with our proposed \acf{WWA} approach.
For this, we pre-trained \acp{BERT} with only \acf{MLM}, relative positional encoding, and \acf{WWM} with the same hyper-parameter setup as in Section \ref{sec:exppre-train}.
Since, as shown in eq.~\eqref{eq:wordtokencomp}, \ac{WWA} allows to increase the maximal token sequence length, we first pre-trained a \ac{BERT} with a token sequence length of 320 (similar to the previous experiments) but also one with a maximum sequence length of 300 words, that is roughly about 450 tokens on average (see eq.~\eqref{eq:wordtokencomp}).
This value is chosen so that approximately the same number of energy values were calculated in this \ac{BERT} model as in the comparable model without \ac{WWA}.

After pre-training of the two \acp{BERT} was finished, we fine-tuned all \ac{NER} tasks with our best fine-tuning method $\text{LCRF}_\text{NER}$.
Table \ref{tab:expwwa} compares the results of using \ac{WWA} to those without (see Table~\ref{tab:expftfr}).

\begin{table}[tbp]
\caption{
Average E-$F_1$ score of three  fine-tuning runs with $\text{LCRF}_\text{NER}$ on \acp{BERT} pre-trained with \acf{MLM}, used relative positional encoding, and \acf{WWM}. Rows refer to the use of \ac{WWA} and the maximal sequence length in pre-training; Columns refer to the use of the Entity-Fix rule.}
\label{tab:expwwa}
\centering
\begin{tabular}{p{1cm}p{3cm}|rr}
\toprule
\textbf{\ac{WWA}} & \textbf{Max seq. length pre-train} & \multicolumn{1}{c}{\textbf{Avg. E-$\bm{F_1}$}} & \multicolumn{1}{p{2.4cm}}{\textbf{Avg. E-$\bm{F_1}$ with Entity-Fix rule}} \\ \midrule
- & 320 token & 0.8651 & 0.8823 \\  
\checkmark & 320 token & 0.8676 & 0.8832 \\ 
\checkmark & 300 words & 0.8674 & 0.8832 \\ 
\bottomrule
\end{tabular}

\end{table}

The results show that, on average, \ac{WWA} slightly improved the original approach, even though training was done with reduced memory consumption due to the smaller number of energy values to be calculated and stored.
The drawback is that the pre-training of a \ac{BERT} with \ac{WWA} takes about 1.5 times longer than pre-training without \ac{WWA} due to the additional window attention layer and the transformation of the sequence from token to word and vice versa. 
The runtime could probably be accelerated by improving the implementation for the transformation from token sequence to word sequence without looping over the samples of the batch.

Increasing the maximum sequence length to 300 words did not result in an additional advantage of the performance.
We suspect that the reason is that the maximum sequence length for \ac{NER} tasks is not a primary concern because the samples in all five \ac{NER} datasets tested almost never exhausted the maximum sequence length used in pre-training.
However, we expect a benefit for other downstream tasks such as document classification or question answering where longer sequences and long-range relations are more important. 

\subsection{Comparing Results with the State of the Art}
In this section, we compare our results with the current state of the art. 
This is only possible for the two \ac{LER} tasks and the GermEval task since the other two \ac{NER} datasets were newly introduced by this paper.

To the best of our knowledge, the current state of the art in both \ac{LER} tasks was achieved in \citep{leitner_fine-grained_2019}.
They applied a non-Transformer model on the basis of bidirectional LSTM layers and a \ac{LCRF} with classical pre-trained word embeddings. 
For evaluation, in contrast to our used E-$F_1$ score, they took a token-wise $F_1$ score, thus, calculating precision and recall per token.

Table \ref{tab:lersota} shows the comparison of the token-wise $F_1$ score to our best results.
\begin{table}[ht]
\caption{Comparison of our best results on the two \ac{LER}-tasks with the previous state of the art models of \protect \cite{leitner_fine-grained_2019}, where T-$F_1$ and E-$F_1$ are the token-, and entity-wise $F_1$ score, respectively.}
\label{tab:lersota}
\centering
\begin{tabular}{p{1.8cm}p{7cm}|rr}
    \toprule
    \textbf{Task} & \textbf{Model} & \multicolumn{1}{c}{\textbf{T-$\bm{F_1}$}} & \multicolumn{1}{c}{\textbf{E-$\bm{F_1}$}}  \\
    \midrule
     \ac{LER} CG & Previous SoTA \citep{leitner_fine-grained_2019} & 0.9595 & - \\ 
     \ac{LER} CG & our best & 0.9842 & 0.9488 \\
     \midrule
     \ac{LER} FG & Previous SoTA \citep{leitner_fine-grained_2019} & 0.9546 & - \\ 
     \ac{LER} FG & our best & 0.9811 & 0.9455 \\
     \bottomrule
\end{tabular}

\end{table}
Note however, that the comparison of the token-wise $F_1$ score is not entirely correct because in our models, words can sometimes break down into several tokens whereas in \citep{leitner_fine-grained_2019} presumably always exactly one token is used per word.
Furthermore, since not published, we were not able to use the identical splits of train, validation, and test data.

Next, in Table~\ref{tab:gesota}, we compare our results for the GermEval task with the current state of the art which, to the best of our knowledge, was achieved in \citep{chan2020german}.
In addition, we list the best results that were achieved without a Transformer architecture \citep{riedl_named_2018}.
\begin{table}[ht]
\caption{Comparison of our best results on the GermEval-task with other state of the art models, where E-$F_1$ is the entity-wise $F_1$ score. Our shown result is again the average of 3 fine-tuning runs. The score of DistilBERT was taken from \url{https://huggingface.co/dbmdz/flair-distilbert-ner-germeval14}}
\label{tab:gesota}
\centering
\begin{tabular}{p{9cm}r|r}
    \toprule
    \textbf{Model} & \multicolumn{1}{c|}{\textbf{Params}} & \multicolumn{1}{c}{\textbf{E-$\bm{F_1}$}}  \\
    \midrule
    BiLSTM-WikiEmb \citep{riedl_named_2018}  & - & 0.8293 \\ 
    $\text{DistilBERT}_{\text{Base}}$ \citep{sanh2019distilbert} & 66\,mio & 0.8563 \\
    $\text{GBERT}_{\text{Base}}$ \citep{chan2020german}  & 110\,mio & 0.8798 \\ 
    $\text{GELECTRA}_{\text{Large}}$ \citep{chan2020german}  & 335\,mio & 0.8895 \\ \midrule
     our best on small \acp{BERT} & 34\,mio & 0.8448 \\
    \bottomrule
\end{tabular}
\end{table}
\begin{table}[ht]
\caption{Comparison of some technical attributes between $\text{GBERT}_{\text{Base}}$ \protect \citep[from][]{chan2020german} and our small \acp{BERT}. The operations per training is a coarse estimation based on the theoretic compute power and pre-training time.}
\label{tab:gehardwcomp}
\centering
\begin{tabular}{l|r|r}
\toprule
 & $\textbf{GBERT}_{\textbf{Base}}$  & \textbf{our small \acp{BERT}} \\ \midrule
Parameter & 110 mio  & 34mio \\ 
Pre-training hardware & 1x TPU v3 & 1x GPU \\ 
Memory & 128GB & 11GB \\ 
Compute power & 420 TFLOPS  & 14 TFLOPS \\ 
Tokens seen in pre-training & $2.6\cdot 10^{11}$  & $1.6\cdot 10^{10}$  \\ 
Pre-training time & $\approx$ 7 days  & $\approx$ 9 days \\ 
Operations per training & $\approx$ 254\,EFLOP & $\approx$ 11\,EFLOP \\ 
\bottomrule
\end{tabular}

\end{table}
While our approach outperforms the BiLSTM results of \cite{riedl_named_2018}, we were not able to reach the results of \cite{chan2020german}. One of the reasons is that our \acp{BERT} are smaller and also pre-trained on a much smaller text corpus. The DistilBERT \citep{sanh2019distilbert} with almost 2x the parameters also reaches a higher score. This comes with the drawback that a larger BERT is needed for pre-training. There are some reasons where it is desired to train a BERT from scratch for example to enable research like the \acl{WWA} or train on a different domain/language.
 In Table~\ref{tab:gehardwcomp} we illustrate some differences of some technical attributes between our \ac{BERT} models and $\text{GBERT}_{\text{Base}}$ of \cite{chan2020german}. It shows that our models can be trained with much lower hardware requirements.

In addition, we compared the time needed for a fine-tuning. 
For a fair comparison, we downloaded the $\text{GBERT}_{\text{Base}}$ from Hugging Face and fine-tuned it with the same hyperparameters as we fine-tuned our models. 
This resulted in an average time of 50 minutes for a fine-tuning run on our models and 70 minutes for a run on  $\text{GBERT}_{\text{Base}}$.

\section{Conclusion and Future Work}
In this article, we conducted our research on comparatively small \ac{BERT} models to address real-world applications with limited hardware resources, making it accessible to a wider audience.
We have worked out how to achieve best results in German \ac{NER} tasks with smaller \ac{BERT} models. 
This simplifies the work of Germanists in the creation of digital editions.

Therefore, we investigated which pre-training method is the most suitable to solve German \ac{NER} tasks on three standard and two newly introduced (H.\,Arendt and Sturm) \ac{NER} datasets.
We examined different pre-training tasks, absolute and relative positional encoding, and masking methods. 
We observed that a \ac{BERT} pre-trained only on the \acf{MLM} task combined with relative positional encoding and \acf{WWM} yielded the overall best results on these downstream tasks.

We also introduced two new fine-tuning variants, $\text{LCRF}_\text{NER}$ and \acf{CSE}, designed for \ac{NER} tasks.
Their investigation in combination with \acl{DFT} and common \acfp{LCRF} showed that the best pre-training technique of the \ac{BERT} is independent of the fine-tuning variant.
Furthermore, we introduced the Entity-Fix rule for decoding.
Our results showed that for most German \ac{NER} tasks, $\text{LCRF}_\text{NER}$ together with the Entity-Fix rule delivers the best results, although there are also tasks for which the \ac{CSE} tagging has a minor advantage.

In addition, our novel \acf{WWA} which modifies the Transformer architecture resulted in small improvements by simultaneously halving the number of energy values to be calculated.
For future work, it would be particularly interesting to investigate \ac{WWA} in connection with other downstream tasks such as document classification or question answering, where the processing of longer sequences is more important than in \ac{NER} tasks.
Another approach would be to combine \ac{WWA} with a sparse-attention mechanism like BigBird \citep{zaheer2020big}.


To further simplify training and application of \acp{BERT} for users with only a low technical background, we are currently developing an  open source implementation of these optimized models in a user friendly software with a graphical user interface.
The goal of this software is to greatly simplify the creation of digital editions by enriching text stored as TEI files with custom \ac{NER} taggings.

\subsection*{Acknowledgments}
This work was funded by the European Social Fund (ESF) and the Ministry of Education, Science, and Culture of Mecklenburg-Western Pomerania (Germany) within the project NEISS (Neural Extraction of Information, Structure, and Symmetry in Images) 
under grant no ESF/14-BM-A55-0006/19.

\section*{Glossary}
\begin{acronym}[MDLSTM]
\acro{BERT}{Bidirectional Encoder Representations from Transformers}
\acro{CSE}{Class-Start-End}
\acro{IOB}{Inside-Outside-Beginning}
\acro{LCRF}{Linear Chain Conditional Random Field}
\acro{LER}{Legal Entity Recognition}
\acro{MLM}{Mask Language Model}
\acro{NER}{Named Entity Recognition}
\acro{NLP}{Natural Language Processing}
\acro{NSP}{Next Sentence Prediction}
\acro{DFT}{Default-Fine-Tuning}
\acro{SOP}{Sentence Order Prediction}
\acro{WWA}{Whole-Word Attention}
\acro{WWM}{Whole-Word Masking}

\end{acronym}

\bibliography{refs} 

\end{document}